\documentclass[runningheads]{llncs}

\usepackage{eccv}

\usepackage{eccvabbrv}

\usepackage{graphicx}
\usepackage{booktabs}
\usepackage{multirow}
\usepackage[accsupp]{axessibility}  

\usepackage{hyperref}

\usepackage{orcidlink}
\usepackage{marvosym}

\begin{document}

\title{Unlocking Attributes' Contribution to Successful Camouflage: A Combined Textual and Visual Analysis Strategy} 

\titlerunning{Unlocking Attributes’ Contribution to Successful
Camouflage}

\author{Hong Zhang\inst{1}\orcidlink{0000-0002-1282-3755} \and
Yixuan Lyu\inst{1}\orcidlink{0000-0001-7608-9150} \and
Qian Yu \inst{1}\orcidlink{0009-0006-7816-2968} \and
Hanyang Liu \inst{1}\orcidlink{0009-0009-8956-1245} \and
Huimin Ma\inst{3}\orcidlink{0000-0001-5383-5667} \and
Ding Yuan\inst{1}\orcidlink{0000-0001-8107-7218} \and
Yifan Yang  \inst{2} \textsuperscript{\Letter} \orcidlink{0000-0003-4237-5874}}

\authorrunning{Zhang \etal}

\institute{School of Astronautics, Beihang University, China \and
Institute of Artificial Intelligence, Beihang University, China 
\email{stephenyoung@buaa.edu.cn}  \and
School of Computer and Communication Engineering, University of Science and Technology Beijing, China}

\maketitle

\begin{abstract}
    In the domain of Camouflaged Object Segmentation (COS), despite continuous improvements in segmentation performance, the underlying mechanisms of effective camouflage remain poorly understood, akin to a black box. To address this gap, we present the first comprehensive study to examine the impact of camouflage attributes on the effectiveness of camouflage patterns, offering a quantitative framework for the evaluation of camouflage designs. To support this analysis, we have compiled the first dataset comprising descriptions of camouflaged objects and their attribute contributions, termed \textbf{COD}-\textbf{T}ext \textbf{A}nd \textbf{X}-attributions (COD-TAX). Moreover, drawing inspiration from the hierarchical process by which humans process information: from high-level textual descriptions of overarching scenarios, through mid-level summaries of local areas, to low-level pixel data for detailed analysis. We have developed a robust framework that combines textual and visual information for the task of COS, named {\bf A}ttribution {\bf CU}e {\bf M}odeling with {\bf E}ye-fixation {\bf N}etwork (ACUMEN). ACUMEN demonstrates superior performance, outperforming nine leading methods across three widely-used datasets. We conclude by highlighting key insights derived from the attributes identified in our study. Code: https://github.com/lyu-yx/ACUMEN.

\end{abstract}

\section{Introduction}
\label{sec:intro}

    Darwinian evolutionary theory posits that wild animals have developed complex camouflage mechanisms to elude predators \cite{stevens2009animal}. Conventional object detection and segmentation algorithms often encounter substantial difficulties in identifying camouflaged objects, resulting in diminished efficacy. This challenge has spurred interest in the field of Camouflaged Object Segmentation (COS), which has seen significant research advancements. Such advancements not only deepen our comprehension of COS but also have practical implications in various fields, including industrial defect detection \cite{fan2023advances}, abnormal tissue segmentation \cite{fan2020pranet}, and the transformation between salient and camouflaged objects \cite{hu2023high}.
    
    Learning in COS has a long history but still striving. Earlier approaches in COS utilized hand-crafted features that were efficient in certain scenarios but fell short in terms of broad applicability \cite{bi2021rethinking}. In contrast, recent studys in data-driven deep learning have shown remarkable success, with techniques leveraging gradients \cite{ji2023deep}, edges \cite{chen2022boundary, lyu2023uedg}, uncertainty \cite{li2021uncertainty}, and multi-view inputs \cite{pang2022zoom}, among others, demonstrating significant improvements. These developments leverage visual features individually but conclude extra priors for guidance, underscore the potential of integrating additional modalities into COS. Very recently, the advent of Large Vision-Language Models (LVLMs) has shifted focus towards harnessing pretrained LVLMs for extracting knowledge, thereby enriching the camouflaged object mask regression process \cite{hu2023relax, cheng2023large}. However, integrating LVLMs directly introduces challenges such as deployment constraints in local environments and costs associated with LVLMs utilization, in addition to the complexities of prompt engineering for COS task.

    Our proposed baseline ACUMEN ({\bf A}ttribution {\bf CU}e {\bf M}odeling with {\bf E}ye-fixation {\bf N}etwork) is underpinned by two critical insights: {\it 1) Cognitive science shows that merging textual and visual information synergistically boosts cognitive understanding} \cite{paivio2013imagery, mayer2002multimedia}, and {\it 2) Evolutionary biology highlights the significance of camouflage pattern creation (by prey) and its identification (by predators) in evolutionary progress, underlining the necessity to analyze camouflage from both granular attribute insights (designing) and a wider object detection (breaking) standpoint}. Capitalizing on the first insight, ACUMEN integrates textual scene descriptions of camouflaged objects. Addressing the second insight, we assess the contribution of potential attributes (e.g., Environmental Pattern Matching, Shape Mimicry) on the efficacy of camouflage. More specifically, we commence by collecting a dataset enriched with image descriptions and attribute contributions. Subsequently, we construct a bifurcated multimodal framework that merges textual and visual analyses seamlessly. Within the textual branch, the framework utilizes frozen CLIP \cite{radford2021learning} text encoder for text analysis, facilitating the synthesis and integration of visual features into a unified latent space. On the visual front, we introduce an attribution and a fixation predictor to assess attribute impacts and generate fixation maps, respectively. Following this prediction phase, an Attributes-Fixation Embedding (AFE) module is implemented to maximize the utility of the predicted attribute contribution tensor and fixation map. This methodology concludes with the delineation of camouflaged objects' masks, accomplished via a transformer decoder and a streamlined projector. Notably, ACUMEN operates solely with the camouflage image during inference, dispensing with the necessity for image descriptions and independence from other LVLMs, thus establishing it as an exclusively visual paradigm.

    To our knowledge, ACUMEN constitutes the first systematic exploration of textual descriptions and attribute contributions within the domain of COS. This investigation uncovers potential for enhancing performance through purely visual methods and provides a deeper understanding of camouflage mechanisms. Our principal contributions are detailed below:

    \begin{itemize}
    \item Introduction of the COD-TAX dataset, which integrates textual information with the COS process.
    \item Preliminary analysis of attribute contributions to camouflage scenes, presenting a novel viewpoint on scene analysis and design.
    \item Development of ACUMEN, a unique dual-branch multimodal fusion framework, setting a new benchmark for cross-modal analysis in the COS field.
    \item Comprehensive experiments demonstrating ACUMEN's superior performance, notably outperforming existing state-of-the-art (SOTA) approaches.
    \end{itemize}

\section{Related works}
    \subsection{Camouflaged Object Segmentation}
    
    The field of COS has consistently garnered interest within the computer vision community. These days, advancements in computing resources, data collection methods, and feature extraction techniques have facilitated a significant shift from traditional handcrafted feature generation \cite{hou2011detection, liu2012foreground, tankus2001convexity} to contemporary data-driven deep learning approaches. Within the deep learning domain, some approaches draw inspiration from biological processes for network architecture design. For example, Fan \etal ~\cite{fan2020camouflaged} emulate the natural predator-prey detection mechanisms, incorporating strategies for search and identification. Similarly, Mei \etal ~\cite{mei2021camouflaged} adopt strategies akin to human positioning and focusing to address COS challenges. Alternatively, certain researchers utilize auxiliary tasks to generate meaningful priors, thereby improving mask regression accuracy. For instance, He \etal ~\cite{he2023camouflaged} employ Wavelet-like feature decomposition and edge detection for supervisory signals, while Lyu \etal ~\cite{lyu2023uedg} leverage uncertainty and edge information for probabilistic and deterministic mask prediction guidance. Wu \etal ~\cite{wu2023source} introduce source-free depth information to enable three-dimensional object analysis. While these methods focus on the manipulation and extraction of visual information, they often overlook the integration of highly condensed semantic supervision and the understanding of camouflage patterns. 
    
    Our proposed ACUMEN model addresses these shortcomings by incorporating textual descriptions as semantic descriptors for high-level consistency supervision and exploring seventeen potential camouflage attributes. This approach not only sets new performance benchmarks on widely used datasets but also provides a deeper insight into various camouflage patterns.

    \subsection{Large Vision-Language Models}
    The surge in interest towards learning from multimodal information, with the aim of achieving coherent representations across varied modalities has marked recent years. This has led to the innovation of various LVLMs designed to bridge the gap between visual and linguistic data. For instance, LLaVA \cite{liu2024visual} combines a vision encoder with a comprehensive language model, enabling detailed interpretations of images based on user instructions. Similarly, BLIP-2 \cite{li2023blip} integrates a pretrained image encoder with a large language model to excel in image-to-text generation tasks. Moreover, Radford \etal \cite{radford2021learning} leveraged 400 million image-text pairs within an end-to-end model known as CLIP, to master open-vocabulary image recognition. This groundbreaking work has catalyzed the creation of numerous applications, including low-light image enhancement \cite{yang2023implicit}, object detection \cite{li2022grounded, zhang2022glipv2}, text-driven image manipulation \cite{lyu2023deltaedit}, and open-vocabulary semantic segmentation \cite{xu2023side, he2023clip}.

    Our proposed methodology seeks to maximize the utilization of the prior knowledge embedded in LVLMs by introducing dual modal branches that extract and maintain coherence between textual and visual information throughout the training phase. This involves the implementation of attributes contribution analysis and fixation prediction mechanisms with the assistance of CLIP, significantly enhance performance.

\section{The COD-TAX Dataset}
    \begin{figure}[tb]
      \centering
      \begin{subfigure}{0.6\linewidth}
      \centering
      \includegraphics[height=5cm]{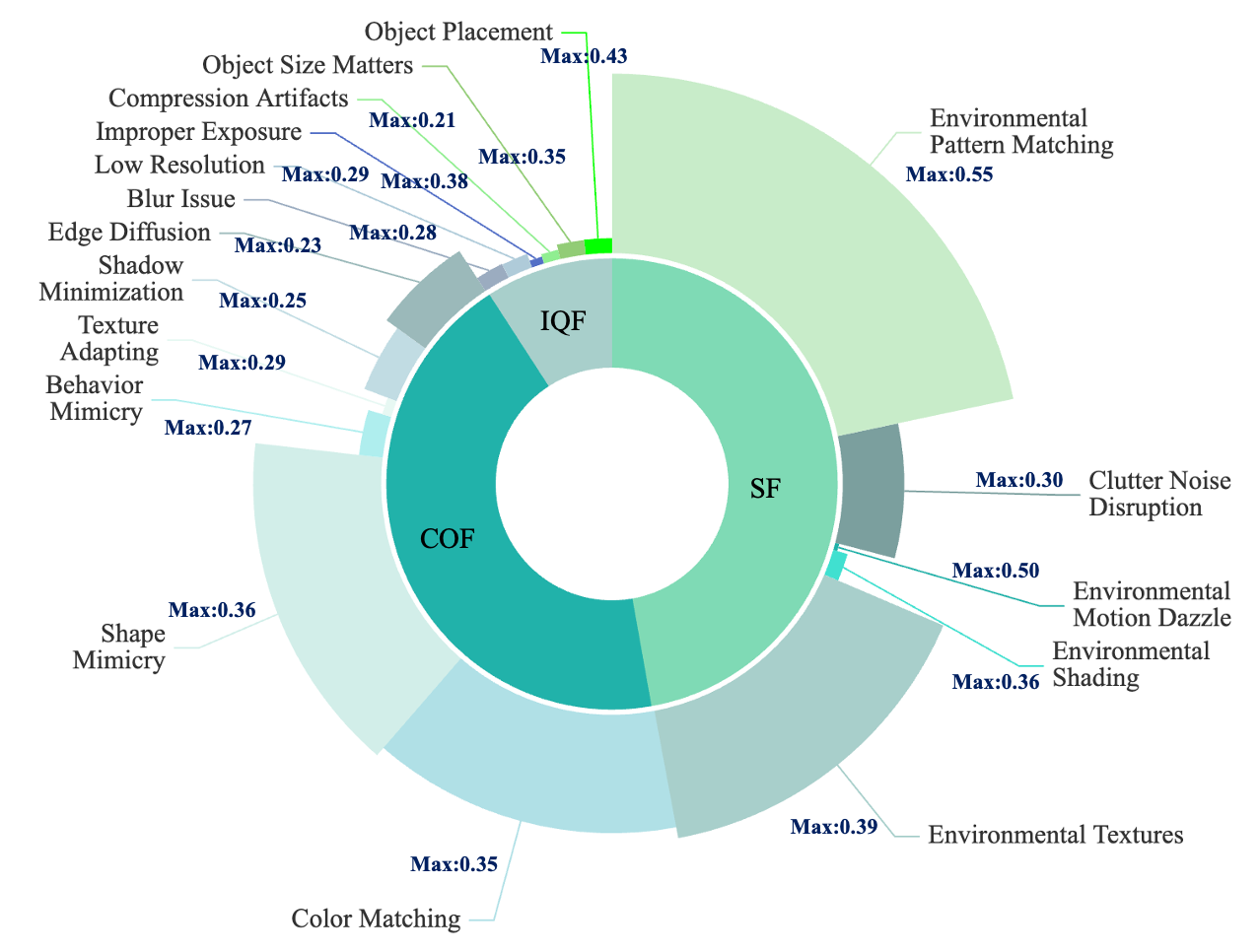}
      \caption{Average attributes' contribution to camouflage}
      \label{fig:distribution_of_contributions}
      \end{subfigure}%
      \hfill
      \begin{minipage}[b]{0.4\textwidth}
        \centering
          \begin{subfigure}{0.5\linewidth}
              \centering
              \includegraphics[height=2cm]{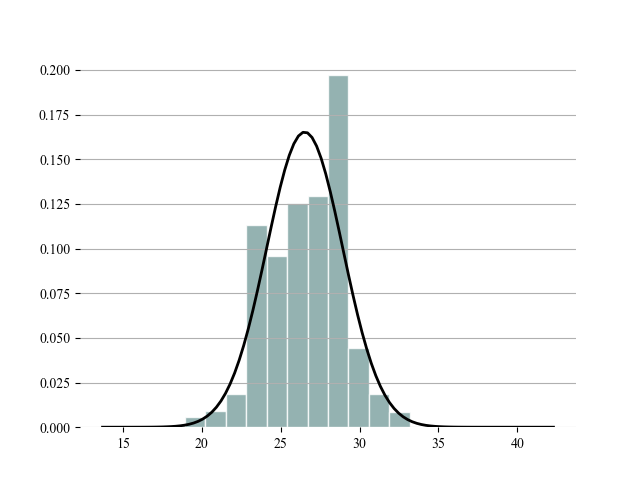}
              \caption{Desc. length}
              \label{fig:word_count}
              \end{subfigure}%
              \hfill
              \begin{subfigure}{0.5\linewidth}
                  \centering
                  \includegraphics[height=2cm]{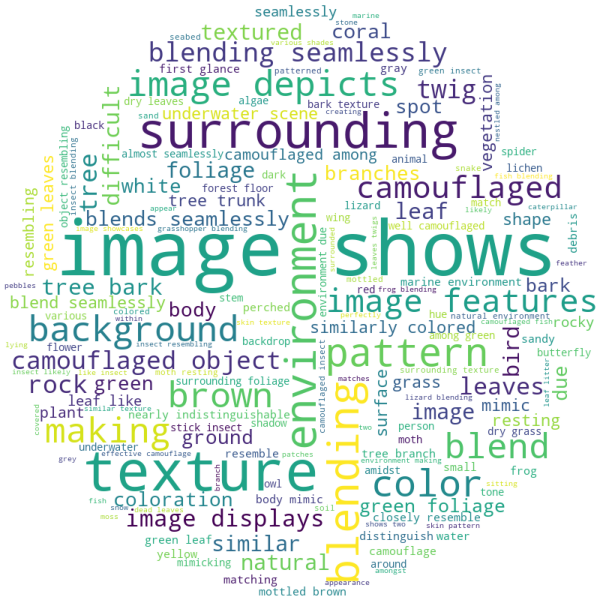}
                  \caption{Word frequency}
                  \label{fig:word_cloud}
              \end{subfigure}%
              \vfill
              \begin{subfigure}{1\linewidth}
                  \centering
                  \includegraphics[height=3cm]{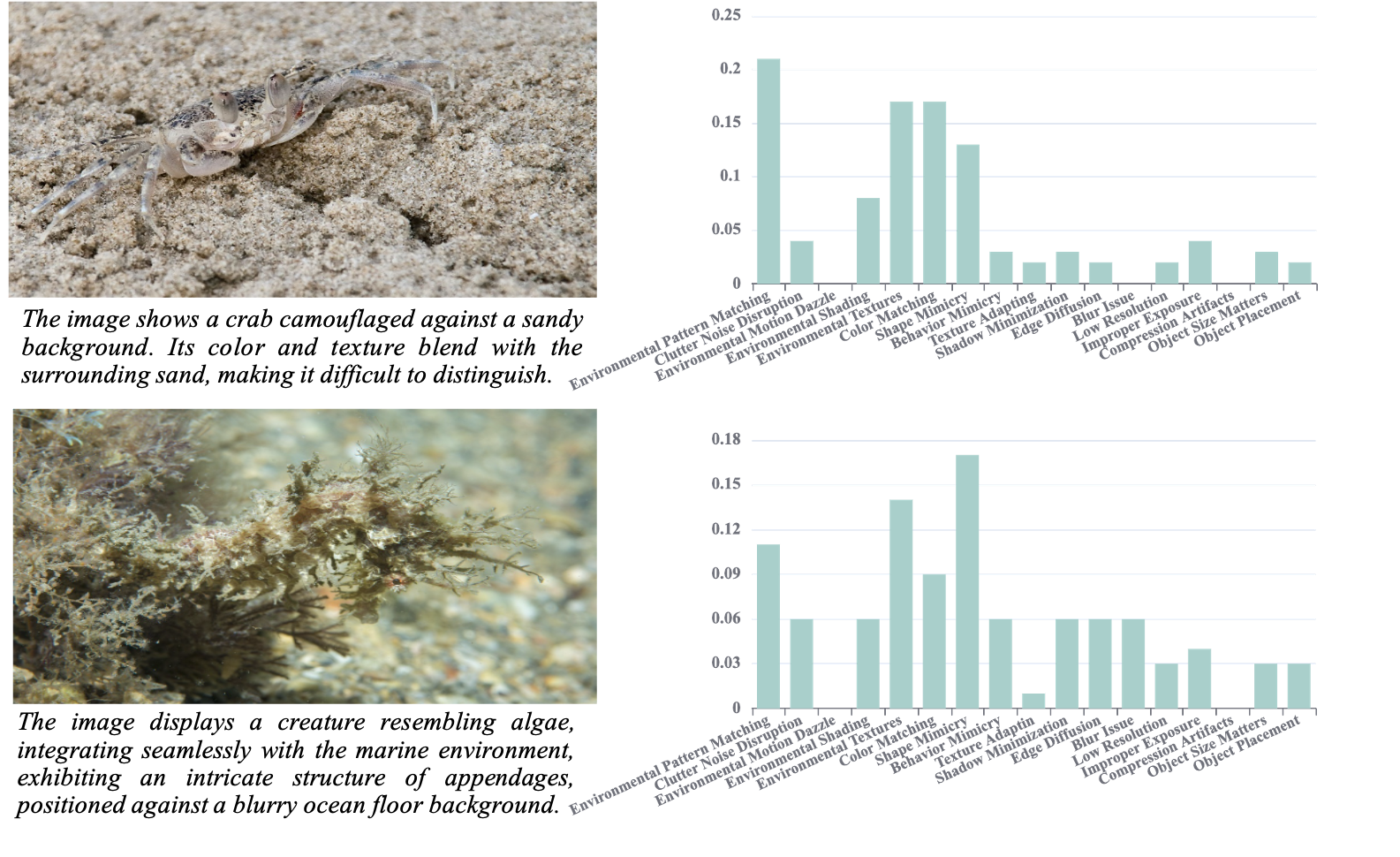}
                  \caption{Illustrations of COD-TAX}
                  \label{fig:data_details}
              \end{subfigure}%
      \end{minipage}
      \caption{Overview of the COD-TAX dataset distribution: (a) 17 attribute classes in three categories, with proportions showing average contributions and \textbf{Max} indicating highest occurrences. (b) Textual description lengths, (c) word cloud of word frequency, and (d) two COD-TAX examples.}
      \label{fig:overall_dataset}
      
    \end{figure}

    \subsection{Text and X-Attributes (TAX) Collecting}
    
    The process of accurately and reliably extracting text descriptions from images containing camouflaged objects is a significant challenge, necessitating extensive time and effort \cite{zhang2023referring}. However, the emergence of LVLMs introduces a promising strategy by exploiting their vast pre-existing knowledge to produce initial descriptions. In this study, we employ GPT-4 Vision (GPT4-V) to generate preliminary descriptions of images and to gain insights into the mechanisms underlying successful camouflage. We predefined a set of potential attributes, allowing GPT4-V to evaluate and determine the importance of each attribute in the observed camouflage. In defining the attributes, we note significant variability in their definitions within the biological domain \cite{merilaita2017camouflage, stevens2009animal, troscianko2009camouflage, pembury2020camouflage, stevens2019key}. Certain attributes, notably "distraction marking" and "internal disruptive", demonstrating considerable overlap, thereby complicating the analysis of camouflage patterns \cite{merilaita2017camouflage}. To address this challenge and broaden the understanding of camouflage-related attributes in the field of computer vision, we expand our categorization methodology based on existing works \cite{merilaita2017camouflage, stevens2009animal, troscianko2009camouflage, pembury2020camouflage, stevens2019key}. Attributes are systematically organized into three primary categories: Surrounding Factors (SF), Camouflaged Object-Self Factors (COF), and Imaging Quality Factors (IQF) as dipicted in \cref{fig:distribution_of_contributions}. This categorization elucidates the origins of camouflage, differentiating between the influences of external environments, inherent characteristics of the camouflaged entity, and constraints imposed by photographic techniques. Each category is extensively detailed, encompassing 17 distinct factors, and a thorough classification is presented in \cref{fig:overall_dataset}.
    
    \subsection{Annotation and Refinement Process}
    To ensure the accuracy and effectiveness of our dataset, we implemented a detailed review process with the participation of over 30 volunteers. These individuals were charged with the critical evaluation of image descriptions produced by GPT4-V and the accuracy of attribute contribution ratios for each image. To enhance the precision of our assessment, we executed three rounds of evaluations for every image. Following the collective insights garnered from these assessments, we accurately identified and amended descriptions and attribute contributions that were consistently deemed incorrect by the evaluators. This rigorous refinement process result in significantly enhanced precision and reliability of the dataset. The comprehensive annotation and refinement effort demanded more than 500 human hours.
    \label{sec:annotation_refinement}

    \subsection{Dataset Features and Statistics}
    We present the statistical analysis of our proposed COD-TAX in \cref{fig:overall_dataset}, offering a comprehensive overview of our dataset. The statistical outcomes, including mean and extreme values, are visualized through a rose chart in \cref{fig:distribution_of_contributions}. In this chart, the size of each petal represents the average contribution value of different attributes under general conditions, highlighting the diverse potential contributions across attributes. The range of maximum values extends from 0.21 to 0.55, whereas the average values fluctuate between 0.004 and 0.21. Additionally, \cref{fig:word_count} provides an analysis of the textual descriptions for each image, with an average length of 26.52 words and a standard deviation of 2.41, indicating that roughly 70\% of descriptions fall within the 24 to 29 word range. In \cref{fig:word_cloud}, we elucidate the frequency of word usage, demonstrating that our dataset predominantly features terms related to surroundings, patterns, backgrounds, textures, and other aspects pertinent to camouflage scene descriptions. To further delineate our dataset's characteristics, we feature two examples in \cref{fig:data_details} to illustrate the distribution proportions of their potential influence attributes, accompanied by detailed image descriptions.

\section{Methods}
    \begin{figure}[tb]
          \centering
          \includegraphics[height=6cm]{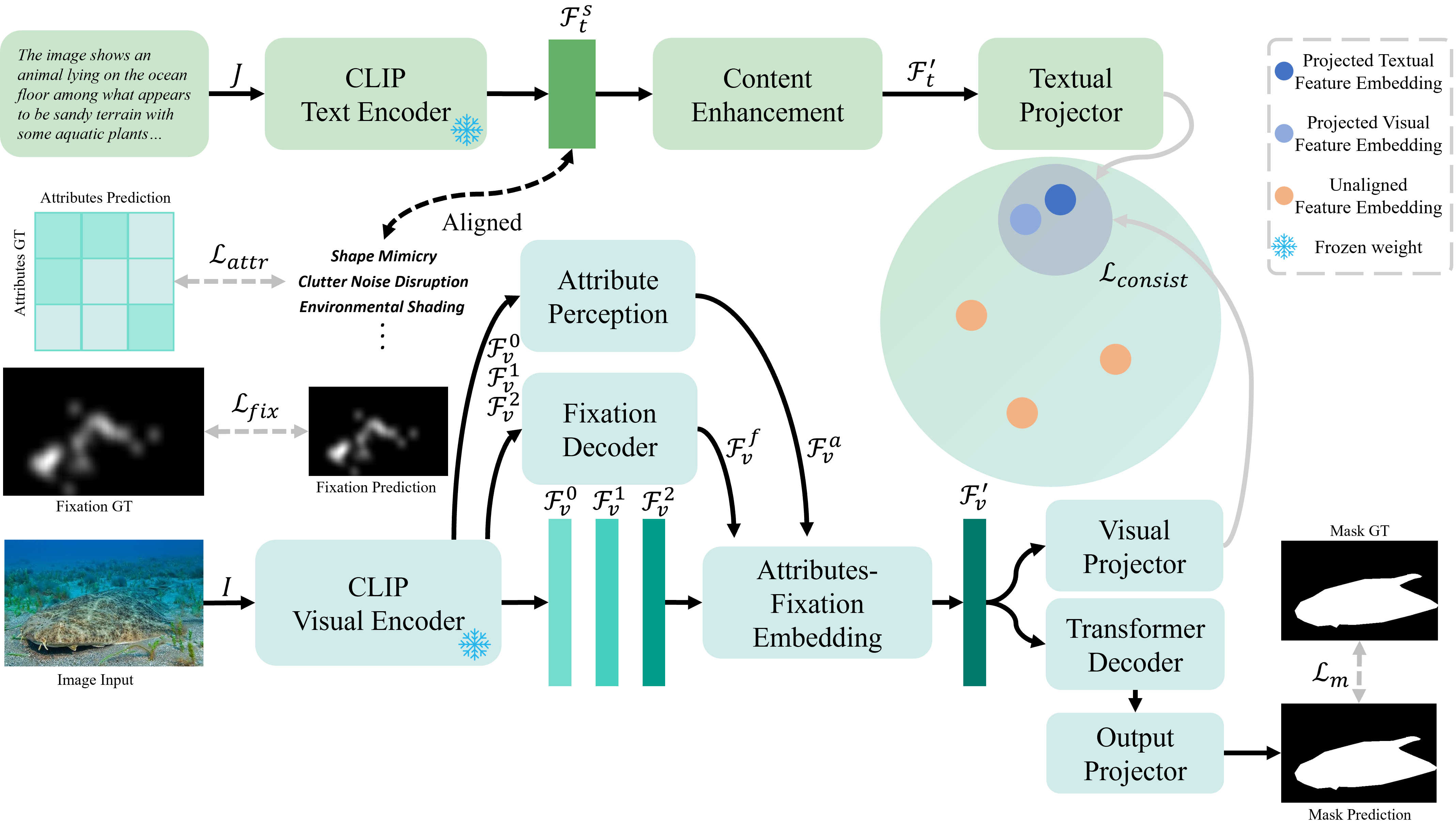}
          \caption{Overall structure of the proposed ACUMEN. The model utilizes both a textual branch and a visual branch, with the textual branch active only during training for practical usage.}
          \label{fig:pipline}
    \end{figure}
    \subsection{Network Overall}
    
    We introduce the comprehensive architecture of the proposed ACUMEN in \cref{fig:pipline}. Initially, we will first detail our underlying motivation and then provide a concise overview of the modules employed in our proposed method.

        {\bf Motivation.}
        Biologically, the evolution of camouflage techniques is significantly influenced by predators' ability to learn and generalize, as well as by prey's behavioral adaptations and decision-making processes aimed at enhancing camouflage effectiveness \cite{CognitionJohn}. However, existing researches predominantly explores the predator's perspective, focusing on developing advanced methods for camouflaged object segmentation. These approaches overlook the prey's strategies, especially those attributes that effectively impair a predator's detection capabilities. To address this imbalance, our framework is designed to not only delineate camouflaged objects but also to evaluate the efficacy of their camouflage attributes through the integration of textual descriptions in COS. This initiative aims to provide a comprehensive abstract representation of camouflage patterns, reflecting both predator and prey dynamics.
        \label{sec:motivation}

        {\bf Network Introduction.}
        As depicted in \cref{fig:pipline}, the ACUMEN framework incorporates a dual-branch architecture, consisting of a textual branch (highlighted in green) and a visual branch (highlighted in cyan), which are pivotal for feature extraction and integration during the training phase. The textual branch, leveraging the CLIP model, processes textual descriptions to distill high-level abstract features, benefiting from the condensing and highly-abstract nature of textual data. Conversely, the visual branch begins by generating human fixation maps to pinpoint mid-level local attention areas, simultaneously predicting the contribution score of attributes. It then leverages these insights for hierarchical embedding, incorporating pixel-level visual features extracted by the CLIP visual encoder. During the inference phase, to enhance the model's applicability, the textual branch is omitted to eliminate dependency on LVLMs like GPT4, thereby making the inference process solely reliant on visual cues.
        \label{sec:network_introduction}

    \subsection{Fixation Prediction}
\begin{figure}[tb]
          \centering
          \includegraphics[height=2.5cm]{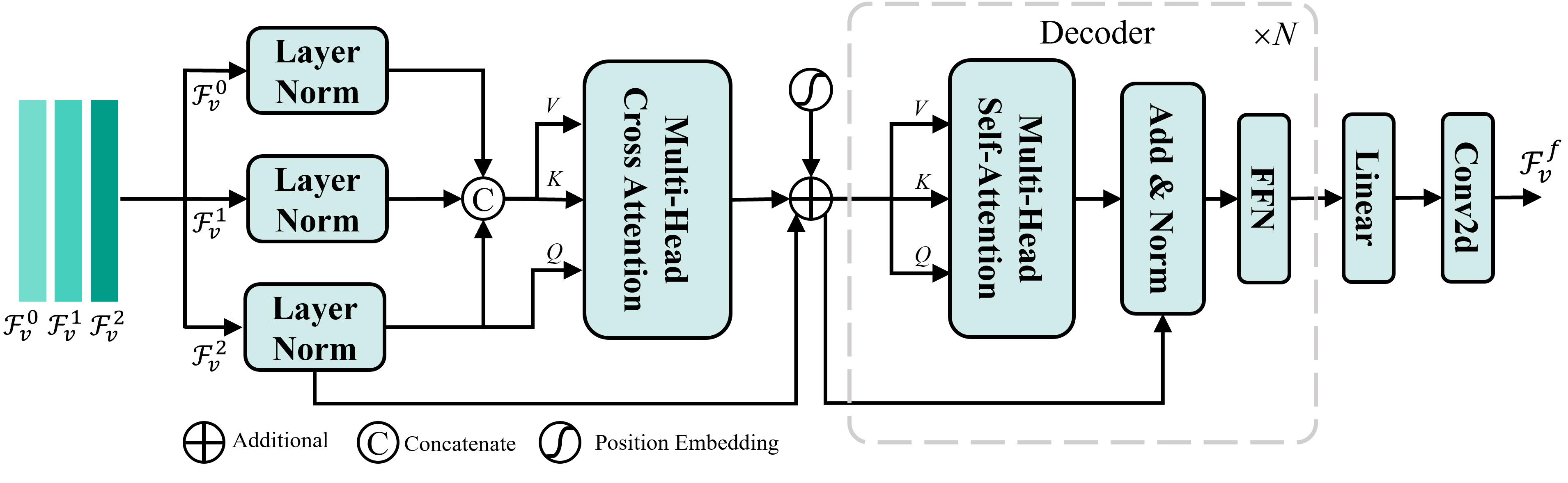}
          \caption{Fixation prediction decoder.
          }
          \label{fig:fix_decoder}
        \end{figure}

    In this study, we employ a Fixation Prediction Module to predict fixations using features from the CLIP visual encoder, as illustrated in \cref{fig:fix_decoder}. Unlike traditional transformer architectures that solely rely on the deepest encoder feature \cite{dosovitskiy2021an}, our approach leverages multiple intermediate features. Specifically, layers 8, 16, and 24 of ViT-L@336, to enhance the information available for fixation prediction tasks. These features are denoted as $F_v^n$ where $n=0, 1, 2$, corresponding to shallow to deep layers. With this strategy, we initially use the deepest vision feature $F_v^2$ as a query to determine its correlation with the concatenated visual features. Subsequently, we adhere to the standard attention mechanism, performing recurrent forward passes $N$ times to produce the final output $F_v^f$ via a linear layer followed by a 2D convolution layer. The fixation prediction process is encapsulated by:
    \begin{equation}
        F_v^f = Conv(Decode(CAtt(F_v^2, Cat[LN(F_v^n)_{n=0, 1, 2}]) + LN(F_v^2) + P_s)_{\times N}).
        \label{eq:fix_prediction}
    \end{equation}
    Here, $P_s$ represents the positional embedding, $LN(\cdot)$ refers to Layer Normalization, $Cat(\cdot)$ indicates channel-wise concatenation, $CAtt$ signifies the Cross-Attention mechanism, and $Decode(\cdot)_{\times N}$ denotes $N$ cascading decoder blocks. For this study, we set $N=3$, as discussed in the ablation study subsection \cref{sec:ablation}. Lastly, $Conv(\cdot)$ represents a sequence of a linear projection followed by a 2D convolution operation.
    For the loss function formulation, the fixation loss is defined as:
    \begin{equation}
        L_{fix} = KL(F_{v}^f, fix_{gt}) + CC(F_{v}^f, fix_{gt}),
    \end{equation}
    where $fix_{gt}$ denotes the ground truth fixation data collected from volunteers~\cite{lv2023towards}. The overall fixation prediction loss is a composite of the Kullback-Leibler (KL) divergence loss and the correlation coefficient (CC) loss, in alignment with standard practices in fixation prediction networks \cite{song2023rinet}.

    \label{sec:fixation_prediction}
    
    \subsection{Attributes' Contribution Prediction}
    We conceptualize the attributes' contribution predicting process as a transformation from high to low dimensions, effectively acting as a dimensionality reduction technique. To achieve this, we employ linear projection complemented by normalization and dropout strategies to enhance training robustness. Specifically, given $F_v^n$ as input, the attribute prediction $F_v^a$ is formally represented as:
    \begin{equation}
        F_v^a = Linear(BRD(Linear(Cat[LN(F_v^n)_{n=0, 1, 2}]))),
    \end{equation}
    Here, $Linear(\cdot)$ refers to linear projection, while $BRD(\cdot)$ signifies the sequential integration of Batch Normalization, ReLU, and Dropout operations.

    To quantify the discrepancy between the actual camouflage attribute contributions and their predictions, the Mean Square Error (MSE) loss is employed for optimization purposes while $attr_{gt}$ indicates the labeled contribute proportion:
    \begin{equation}
        L_{attr} = MSE(F_v^a, attr_{gt}).
    \end{equation}
    \label{sec:attribute_prediction}
    
    \subsection{Attributes-Fixation Embedding}
    \begin{figure}[tb]
      \centering
      \includegraphics[height=3cm]{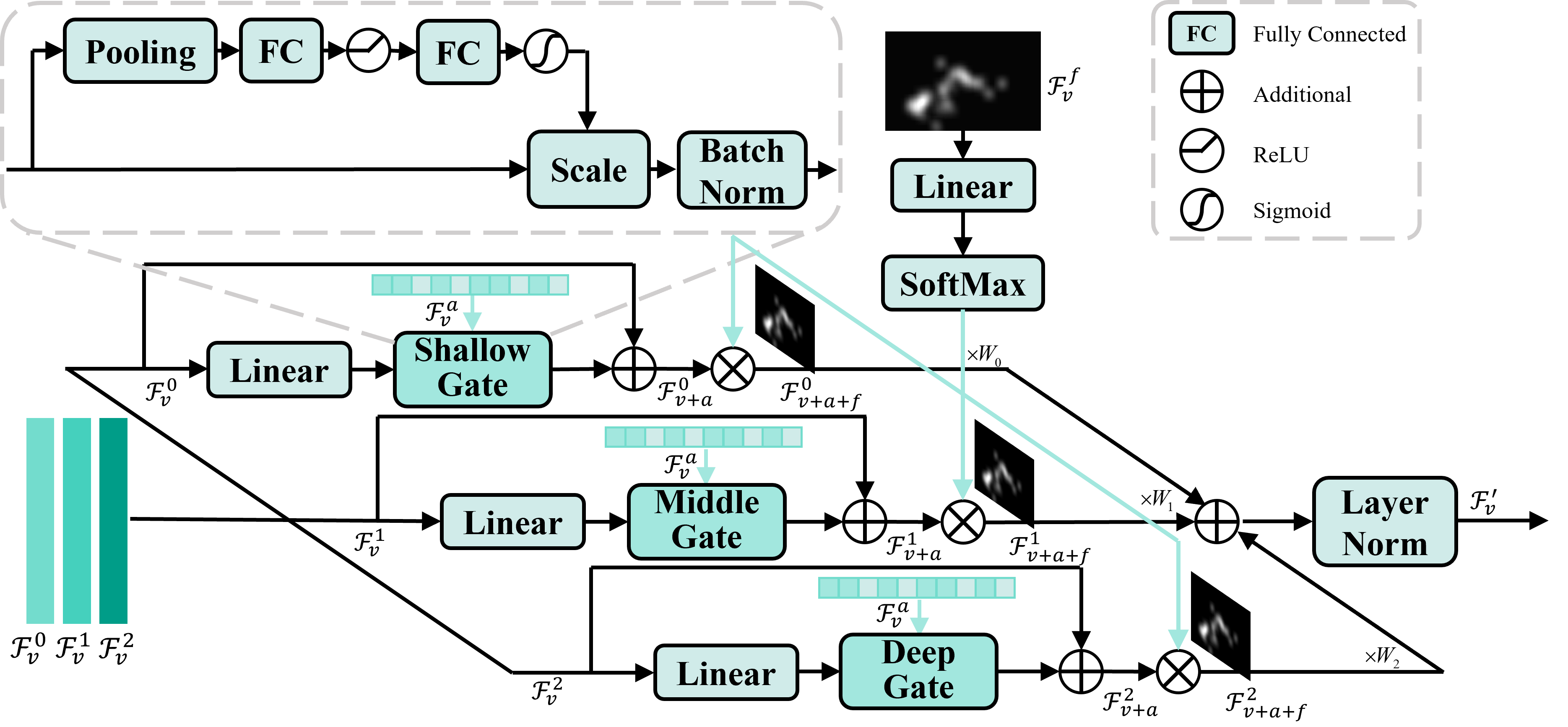}
      \caption{Attributes-Fixation Embedding structure. }
      \label{fig:AFE}
    \end{figure}

    To effectively leverage attribute information and fixation maps obtained from the fixation and attribute decoders, we introduce the Attributes-Fixation Embedding (AFE) approach as depicted in \cref{fig:AFE}. This method incorporates these elements as supplementary priors alongside raw CLIP visual features. Specifically, the visual features are processed through three distinct branches. In each branch, following the linear projection, the resulting features are directed to their respective gating mechanisms. Acknowledging the potential interrelations among attributes and the necessity for channel-wise feature recalibration, we adopt the Squeeze-and-Excitation (SE) mechanism \cite{hu2018squeeze} to facilitate the fusion of attribute information. Subsequently, the fixation map $F_v^{f}$ is employed as a biologically interpretable attention mechanism to augment the features within each branch. Furthermore, to prioritize the features from deeper ViT layers, which exhibit finer granularity after successive layers, weights are assigned to each branch prior to summation. Ultimately, the AFE feature $F_v'$ is generated, followed by a Layer Normalization operation. The entire AFE process is formalized as follows:
    \begin{align}
        F_{v+a}^i &= Gate(Linear(F_{v}^i), F_{v}^a) + F_{v}^i, \\
        F_{v+a+f}^i &= Mul(F_{v+a}^i, Softmax(Linear(F_v^f))), \\
        F_v' &= LN(\frac{1}{M}\sum_{i=0}^{2}Mul(W_i, F_{v+a+f})). 
    \end{align}
    Here, $i$ denotes the $i$-th branch and $F_{v+a}^i$ represents the feature enhanced with attribute information, $F_{v+a+f}^i$ signifies the feature further refined by the fixation map, and $F_v'$ is the outcome of the AFE process. The $Gate(\cdot)$ operation employs the SE mechanism to integrate attribute information into the visual feature. $Mul(\cdot)$ denotes element-wise multiplication, and $\sum(\cdot)$ signifies element-wise addition. The weights $W_i$ are assigned with values of 1, 2, and 4 for $W_0$ to $W_2$ respectively, and $M=\sum_{i=0}^{2}W_i$ serves as the normalization constant.
    
    \subsection{Mask Predicting}
    After acquiring the visual feature with embedded camouflage attribute and fixation information, represented as $F_{v}'$, we proceed to utilize a general transformer decoder and an output projector to unveil the final camouflage object mask $M_{p}$. The unveiling process of this mask can be formulated as:
    \begin{equation}
        M_{p} = Conv2d(CBR(Decoder(F_{v}')_{\times M})_{up_4}),
    \end{equation}
    
    where $Conv2d(\cdot)$ denotes the 2D convolution, and $CBR(\cdot)$ signifies the sequential layers of Convolution, Batch Normalization, and ReLU. $Decoder(\cdot)_{\times M}$ represents the transformer decoder executed with $M$ iterations, discussed in \cref{sec:ablation}. The subscript $up_4$ indicates the 4 times upsampling operation.
    
    Moreover, the loss function $L_{mask}$ is constructed from the weighted binary cross entropy (wBCE) loss and the weighted Intersection over Union (wIoU) loss, adhering to conventional practices \cite{wei2020f3net, lyu2023uedg}:
    \begin{equation}
        L_{mask} = L_{BCE}^w + L_{IoU}^w
    \end{equation}

    \subsection{Total Loss Function}
    To enhance the use of high-level, condensed textual information obtained from the CLIP text encoder, we propose a novel consistency measurement mechanism aimed at monitoring the manipulation of visual features throughout the training phase. We have developed two distinct projectors to map both the overall description feature, denoted as $F_t'$, and the refined visual feature, denoted as $F_v'$, into a unified latent feature space. The feature $F_t'$ is obtained from the output of the CLIP text encoder, $F_t^s$. Considering that both features relate to the identical camouflage image, they should demonstrate consistency in this latent space. To measure this consistency, we employ a consistency loss, $L_{consist}$, defined as:
    \begin{equation}
        L_{consist} = CS(Proj(F_v')_v, Proj(F_t')_t),
    \end{equation}
    where $CS(\cdot)$ denotes the cosine similarity loss. $Proj(\cdot)_t$ and $Proj(\cdot)_v$ represent the projectors for mapping into the latent feature space. The total loss function is formulated as follows, where $\alpha, \beta,$ and $\gamma$ serve as the balancing weights:
    \begin{equation}
        L_{total} = L_{mask} + \alpha L_{fix} + \beta L_{attr} + \gamma L_{consist}.
    \end{equation}
\section{Experiments}
    \subsection{Experimental Settings}
    \textbf{Dataset:} In accordance with the protocol \cite{fan2021concealed} and building upon the experimental framework \cite{lyu2023uedg}, we employ a combined dataset for training, comprising CAMO-train\cite{le2019anabranch} and COD10K-train\cite{fan2021concealed}, totaling 4040 images. For evaluation, we utilize the CAMO, COD10K, and NC4K datasets\cite{lv2021simultaneously}, containing 250, 2026, and 4121 images, respectively.
    
    \textbf{Evaluation Metrics:} The evaluation of network performance during training utilizes four widely recognized metrics: structure-measure ($S_\alpha$)\cite{fan2017structure}, weighted F-measure ($F_\beta^{\omega}$)\cite{margolin2014evaluate}, mean enhanced-alignment measure ($E_{\phi}$)\cite{fan2018enhanced}, and mean absolute error ($M$).
    
    \textbf{Implementation Details:}
    The training and testing procedures are conducted using PyTorch on NVIDIA RTX 8000 GPUs. We uniformly resize input images to $336\times336$ pixels to comply with the requirements of the pretrained CLIP model, ViT-L@336. The optimization process employs the Adam algorithm\cite{kingma2014adam}, accompanied by a multi-step learning rate schedule. The initial learning rate is established at $1e^{-4}$, with a decay factor of 0.2 applied following 150 epochs. Completing the training regimen over 200 epochs takes approximately 16 hours on four NVIDIA RTX 8000 GPUs.
    
    \subsection{Comparing with SOTA methods}

\begin{figure}[tb]
              \centering
              \includegraphics[height=4.5cm]{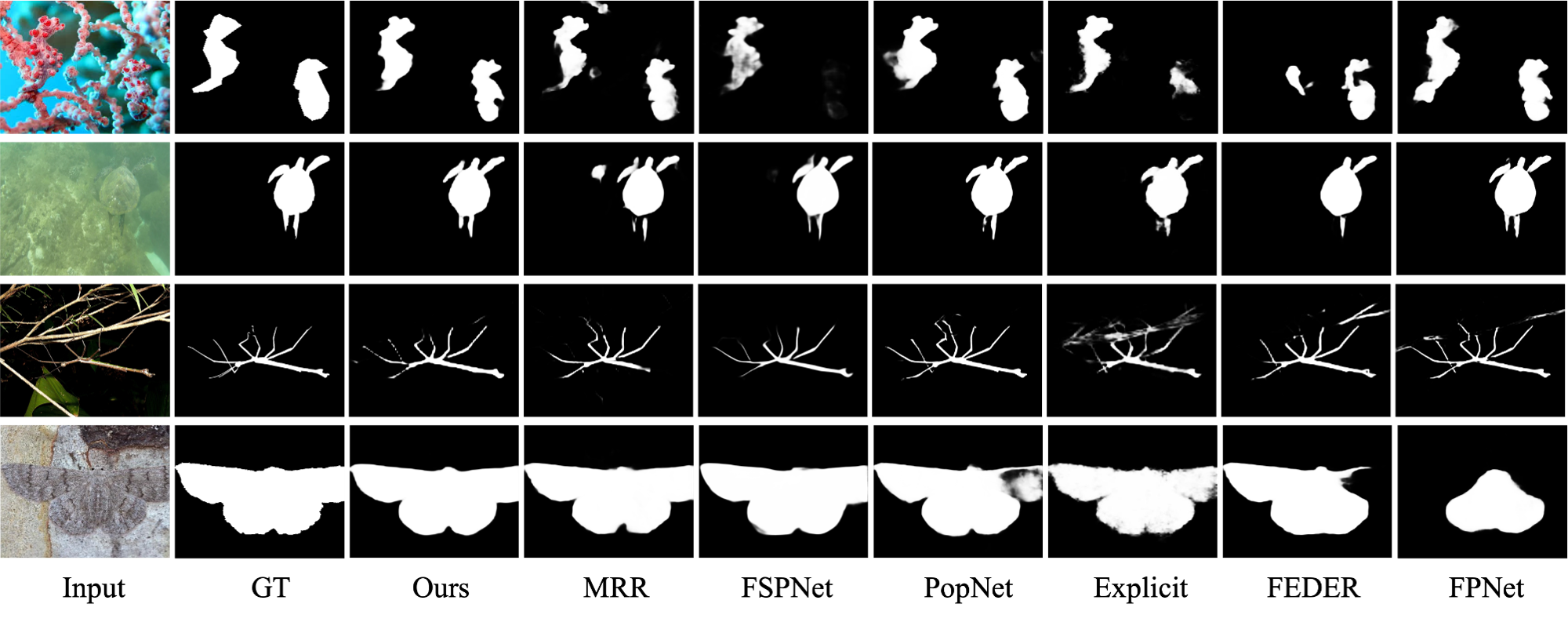}
              \caption{Qualitative comparison of ACUMEN with SOTA methods.
              }
              \label{fig:Qualitative_analysis}
            \end{figure}

        \textbf{Qualitative Results.} 
            The results presented in Figure~\ref{fig:Qualitative_analysis} highlight the superior predictive performance of our proposed ACUMEN, when compared to SOTA methods in a variety of scenarios. Notably, in underwater settings as shown in the top row of the figure, other methods often fail to accurately identify two mimicry seahorses, suffering from issues like partial detection, edge blurring, and incorrect counts. In contrast, ACUMEN achieves remarkable visual clarity, leading to more accurate and comprehensive predictions. Furthermore, in terrestrial environments, illustrated in the third row, our method excels at precisely detecting the fine limbs of the stick insect, distinguishing them from dead branches without the issues of blurring or ambiguity. This enhanced ability to discern the structure of objects is primarily attributed to the integration of high-level textual information, introducing object structure priors into the supervision of the visualization feature extraction process.

        {\bf Quantitative Results.}
            The effectiveness of ACUMEN was evaluated by comparison with SOTA methods across three datasets using four evaluation metrics. As shown in \cref{tab:quant_result}, ACUMEN consistently surpasses competing methods in all datasets, demonstrating significantly superior performance. Specifically, within the CAMO dataset, ACUMEN significantly outshines other methods in all metrics, with $S_\alpha$ and $F_\beta^{\omega}$ scores of 0.886 and 0.850, respectively, which are 3.5\% and 5.5\% better than the second best method. In the NC4K dataset, although ACUMEN's $S_\alpha$ and $M$ scores are on par with FSPNet \cite{huang2023feature}, its $E_{\phi}$ and $F_{\beta}^w$ scores are notably higher. Importantly, ACUMEN achieves SOTA performance while utilizing the smallest input size among its counterparts. Considering the significant advantages that larger inputs can provide in enhancing COS performance\cite{hu2023high}, ACUMEN shows potential for even greater effectiveness with equivalent input resolutions.

    \subsection{Ablation Study}
        \begin{table}[tb]

            \caption{Quantitative comparison of our method with nine SOTA approaches on three benchmark datasets, with the highest scores highlighted in \textbf{bold}. To ensure consistency in evaluation metrics, we recalculated the results of methods with available resources using the evaluation protocols defined in \cite{fan2020camouflaged}. Metrics not originally reported are marked with an asterisk "*" to denote these recalculations. For methods without publicly available results, we denote their absence with "{$\dag$}".}

            \label{tab:quant_result}
            \centering
            \resizebox{\textwidth}{!}{
                \fontsize{4}{12}\selectfont 
                \begin{tabular}{c|c|c|cccc|cccc|cccc}
                \toprule
                \multicolumn{1}{c|}{\multirow{2}{*}{Methods}} & \multicolumn{1}{c|}{\multirow{2}{*}{Publication}} & \multirow{2}{*}{Size} & \multicolumn{4}{c|}{CAMO} & \multicolumn{4}{c|}{COD10K} &
                \multicolumn{4}{c}{NC4K} \\ \cline{4-15}
                \multicolumn{1}{c|}{} & \multicolumn{1}{c|}{} & &  $S_\alpha \uparrow$ & $E_{\phi} \uparrow$ & $F_\beta^{\omega} \uparrow$ & $M \downarrow$ &  \multicolumn{1}{c}{}
                $S_\alpha \uparrow$ & $E_{\phi} \uparrow$ & $F_\beta^{\omega} \uparrow$ & $M \downarrow$
                & \multicolumn{1}{c}{}
                $S_\alpha \uparrow$ & $E_{\phi} \uparrow$ & $F_\beta^{\omega} \uparrow$ & $M \downarrow$ \\ \midrule
                PopNet\cite{wu2023source} & ICCV$_{\text{2023}}$ 
                & $512^2$ & 0.806 & $0.859^*$ & $0.744^*$ & 0.073 & 0.827 & $0.910^*$ & $0.757^*$ & 0.031 & 0.852 & $0.909^*$ & $0.802^*$ & 0.043 \\
                CFANet\cite{zhang2023cfanet} & ICME$_{\text{2023}}$ & $416^2$ & 0.815 & 0.876 & 0.761 & 0.073 & 0.834 & 0.905 & 0.730 & 0.031 & 0.848 & 0.906 & 0.791 & 0.046 \\
                MFFN\cite{zheng2023mffn} & WACV$_{\text{2023}}$ & $384^2$ & {$\dag$} & {$\dag$} & {$\dag$} & {$\dag$} & 0.846 &  $0.897^*$ & 0.745 & 0.028 & 0.856 & $0.902^*$ & 0.791 & 0.042 \\
                FEDER\cite{he2023camouflaged} & CVPR$_{\text{2023}}$ 
                & $384^2$ & 0.807 & 0.873 & $0.738^*$ & 0.069 & 0.823 & 0.900 & $0.716^*$ & 0.032 & 0.846 & 0.905 & $0.789^*$ & 0.045 \\
                Explicit\cite{liu2023explicit} & CVPR$_{\text{2023}}$ & $352^2$ & 0.846 & 0.895 & 0.777 & 0.059 & 0.843 & 0.907 & 0.742 & 0.029 & {$\dag$} & {$\dag$} & {$\dag$} & {$\dag$} \\
                FSPNet\cite{huang2023feature} & CVPR$_{\text{2023}}$ 
                & $384^2$ & 0.856 & 0.899 & 0.799 & 0.050 & 0.851 & 0.895 & 0.735 & 0.026 & \textbf{0.879} & 0.915 & 0.816 & \textbf{0.035} \\
                MRR-Net\cite{yan2023camouflaged} & TNNLS$_{\text{2023}}$ & $384^2$ & 0.826 & 0.880 & $0.759^*$ & 0.070 & 0.835 & 0.901 & $0.720^*$ & 0.032 & 0.857 & 0.906 & $0.786^*$ & 0.044 \\
                FPNet\cite{cong2023frequency} & ACM MM$_{\text{2023}}$ & $512^2$ & 0.852 & 0.905 & 0.806 & 0.056 & 0.850 & 0.913 & 0.748 & 0.029 &{$\dag$} & {$\dag$} & {$\dag$} & {$\dag$} \\
               $\text{LSR+}^2$\cite{lv2023towards} & TCSVT$_{\text{2023}}$ & $384^2$ & 0.854 & 0.924 & {$\dag$} & 0.049 & 0.847 & 0.924 & {$\dag$} & 0.028 & 0.870 & 0.924 & {$\dag$} & 0.036 \\
                
                \midrule 
                Ours & - & $336^2$ & \textbf{0.886} & \textbf{0.939} & \textbf{0.850} & \textbf{0.039} & \textbf{0.852} & \textbf{0.930} & \textbf{0.761} & \textbf{0.026} & 0.874 & \textbf{0.932} & \textbf{0.826} & 0.036 \\ 
                \bottomrule
                \end{tabular}}
        \end{table}
        \label{sec:ablation}


        \begin{table}[]
        \caption{Hyper parameter Search. The best result are \bf{bold}.}
        \label{tab:meta_parameter}
        \centering
            \fontsize{3}{7}\selectfont 
            \begin{tabular}{ccc|cc|cccc|cccc}
            \toprule
            \multirow{2}{*}{$N$} & \multirow{2}{*}{$M$} & \multirow{2}{*}{$WL$} & \multirow{2}{*}{Gmac} & \multirow{2}{*}{Params} & \multicolumn{4}{c|}{CAMO}                                                                  & \multicolumn{4}{c}{COD10K}                                                                 \\ \cline{6-13} 
                                 &                      &                       &                       &                         & $S_\alpha \uparrow$ & $E_{\phi} \uparrow$ & $F_\beta^{\omega} \uparrow$ & $M \downarrow$   & $S_\alpha \uparrow$ & $E_{\phi} \uparrow$ & $F_\beta^{\omega} \uparrow$ & $M \downarrow$   \\ \midrule
            1                    & 1                    & 50                    & 259.07                & 484.65                  & 0.87586             & 0.93248             & 0.83537                     & 0.04139          & 0.84668             & 0.92401             & 0.7509                      & 0.02701          \\
            1                    & 1                    & 77                    & 259.07                & 484.65                  & 0.88475             & 0.93505             & 0.83766                     & 0.04044          & 0.85105             & 0.93083             & 0.7467                      & 0.02740           \\
            1                    & 3                    & 50                    & 262.71                & 490.96                  & 0.88054             & 0.93567             & 0.84328                     & 0.03972          & 0.84876             & 0.93100               & 0.75671                     & 0.02652          \\
            1                    & 3                    & 77                    & 262.71                & 490.96                  & 0.88061             & 0.93128             & 0.83028                     & 0.04183          & 0.85038             & 0.92851             & 0.74543                     & 0.02654          \\
            1                    & 6                    & 50                    & 268.16                & 500.41                  & 0.86630              & 0.91529             & 0.80757                     & 0.04932          & 0.83619             & 0.91431             & 0.72194                     & 0.03249          \\
            1                    & 6                    & 77                    & 268.16                & 500.42                  & 0.88049             & 0.93489             & 0.83709                     & 0.04044          & 0.84975             & 0.93212             & 0.75163                     & 0.02681          \\
            3                    & 1                    & 50                    & 262.71                & 490.96                  & 0.88568             & 0.93894             & \textbf{0.8496}             & \textbf{0.03879} & 0.85153             & 0.93035    & \textbf{0.76149}            & 0.02592          \\
            3                    & 1                    & 77                    & 262.71                & 490.96                  & 0.88257             & 0.93752             & 0.84623                     & 0.03907          & 0.85170             & \textbf{0.9337}              & 0.76056                     & \textbf{0.02579} \\
            3                    & 3                    & 50                    & 266.34                & 497.27                  & \textbf{0.88803}    & 0.93375             & 0.8399                      & 0.04079          & \bf{0.85311}        & 0.93021             & 0.74850                     & 0.02699          \\
            3                    & 3                    & 77                    & 266.34                & 497.27                  & 0.88177             & 0.93358             & 0.84139                     & 0.04070           & 0.85051             & 0.93026             & 0.75614                     & 0.02667          \\
            3                    & 6                    & 50                    & 271.80                & 506.72                  & 0.88389             & 0.93615             & 0.84137                     & 0.03964          & 0.84948             & 0.93265             & 0.74997                     & 0.02676          \\
            3                    & 6                    & 77                    & 271.80                & 506.72                  & 0.88552             & \textbf{0.94005}    & 0.8474                      & 0.03892          & 0.84948             & 0.93052             & 0.74997                     & 0.02676          \\ \bottomrule
            \end{tabular}
            \end{table}

        \textbf{Number of Decoder Layers.} The discussion of hyperparameters is detailed in \cref{tab:meta_parameter}. Here, $N$ and $M$ denote the number of decoder layers in the fixation and mask prediction modules, respectively. Additionally, $WL$ is the word length of the input description, with a default value of 77 in the CLIP textual encoder. Our analysis indicates that an $N$ value of 3 and an $M$ value of 1 substantially improve the performance of $F_\beta^{\omega}$ over other configurations, while ensuring similar levels of performance metrics $E_{\phi}$, $S_\alpha$, and $M$, as evidenced by the CAMO and COD10K datasets in \cref{tab:meta_parameter}. Moreover, our review of sentence lengths, as depicted in \cref{fig:word_count}, confirms that they do not exceed 50 words. This observation supports our decision to reduce the default word length from 77 to minimize blank input.

        \begin{table}
            \caption{ACUMEN components' contribution discussion. The best result are \bf{bold}.}
            \label{tab:module_contribution}
            \centering
                \fontsize{4}{7}\selectfont 
                \begin{tabular}{ccc|cc|cccc|c}
                \toprule
                \multirow{2}{*}{Fix} & \multirow{2}{*}{Attr} & \multirow{2}{*}{Const} & \multirow{2}{*}{Gmac} & \multirow{2}{*}{Params} & \multicolumn{4}{c|}{COD10K} & \multirow{2}{*}{FPS} \\ \cline{6-9}
                 &  &  &  &  & $S_\alpha \uparrow$ & $E_{\phi} \uparrow$ & $F_\beta^{\omega} \uparrow$ & $M \downarrow$ &  \\ \midrule
                 &  &  & 255.13 & 227.99 & 0.84412 & 0.92106 & 0.72491 & 0.02945 & \bf{14.18}\\
               \checkmark  &  &  & 262.36 & 264.13 & 0.84452 & 0.92446 & 0.74769 & 0.02769 &  12.30\\
                 & \checkmark &  & 256.74 & 480.61 & 0.85132 & 0.92692 & 0.75853 & 0.02653  & 13.35\\
               \checkmark  & \checkmark & & 262.71 & 490.96 & 0.84972 & 0.92838 & 0.75797 & 0.02626  & 12.19 \\
               \checkmark  & \checkmark & \checkmark & 262.71 & 490.96 & \bf{0.85153} & \bf{0.93035} & \bf{0.76149} & \bf{0.02592} & 12.20\\ \bottomrule
                \end{tabular}
        \end{table}
        \textbf{Contribution of Component Modules.} The contributions of individual module are summerized in \cref{tab:module_contribution}. Here, "Fix" corresponds to the process of fixation prediction, while "Attr" signifies the attribute prediction process. The term "Const" is used to describe the mechanism that ensures coherence between visual and textual features. Our analysis indicates that the exclusion of any module results in a degradation of the ACUMEN model's performance. Importantly, incorporating the attribute prediction mechanism significantly increases the model's parameter count, mainly due to the addition of the CLIP textual encoder. The consistency mechanism, which is only applied during the training phase for supervisory purposes, does not affect the Gmac or parameter count, thereby ensuring uniformity in our experiments. Additionally, there is only a negligible decrease in Frames Per Second (FPS), which can be attributed to the slight increment in Gmac.
    \subsection{Intermediate Outputs Visualization}

    \begin{figure}[tb]
      \centering
      \begin{subfigure}{0.5\linewidth}
        \centering
        \includegraphics[height=4cm]{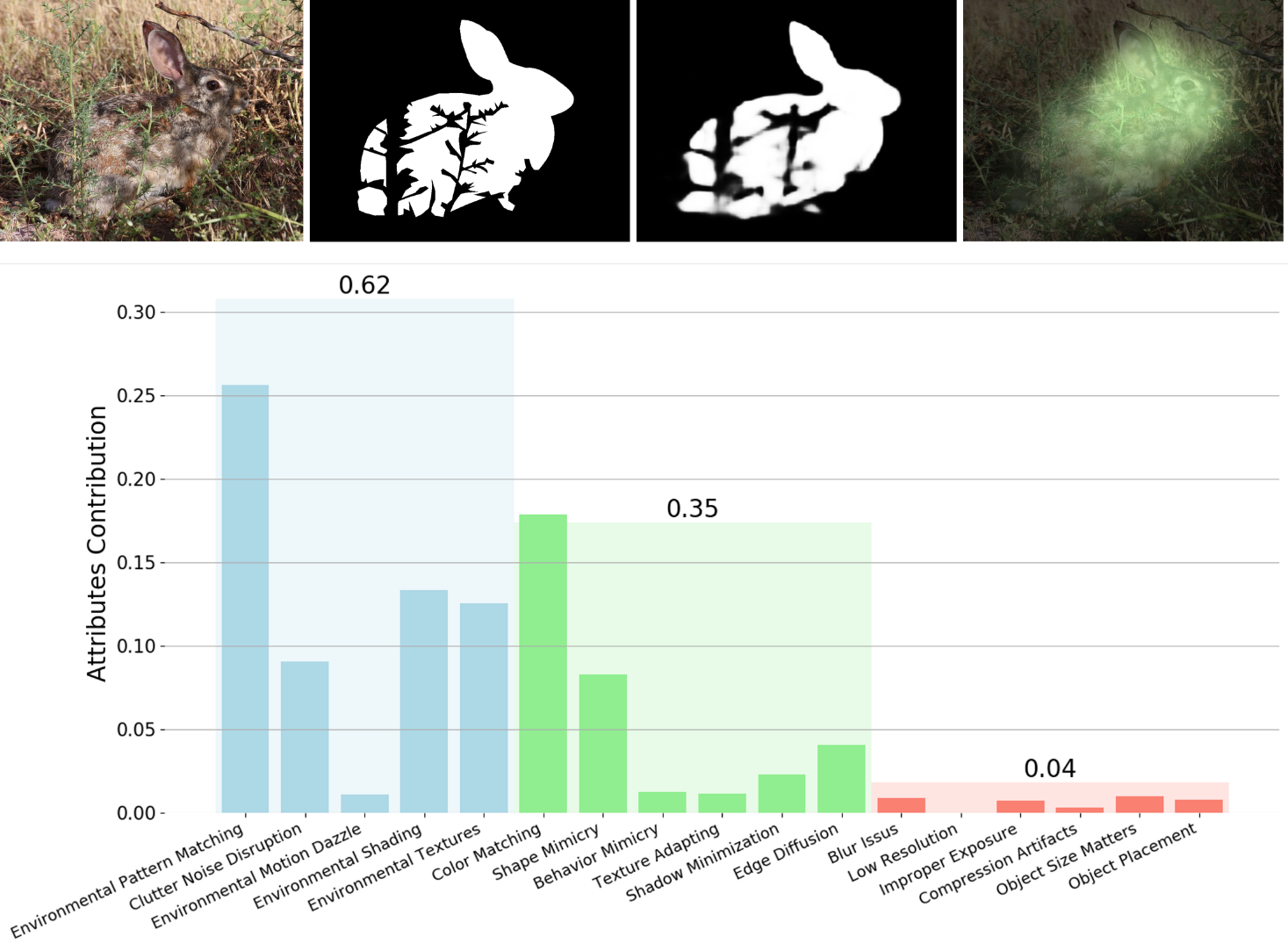}
        \caption{Surrounding Factors dominate}
        \label{fig:intermediate_output_a}
      \end{subfigure}%
      \hfill
      \begin{subfigure}{0.5\linewidth}
        \centering
        \includegraphics[height=4cm]{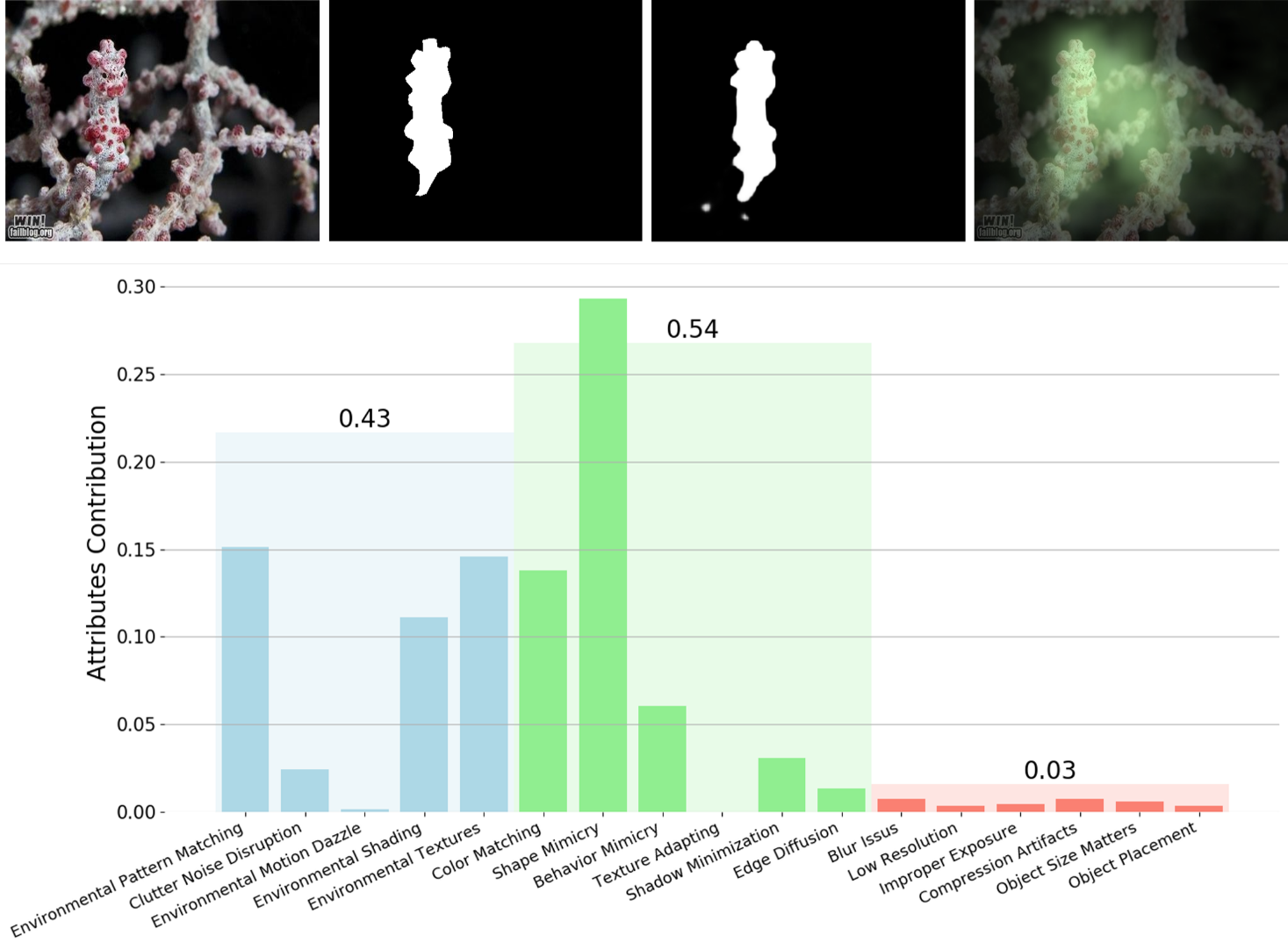}
        \caption{Camouflaged Object-Self Factors dominate}
        \label{fig:intermediate_output_b}
      \end{subfigure}
      \caption{Examples of intermediate outputs}
      \label{fig:intermediate_output}
    \end{figure}
  
    To elucidate the effectiveness of ACUMEN's components, we demonstrate the final and intermediate outputs in \cref{fig:intermediate_output}. For each example, the top row illustrates the input image, ground truth mask, our generated result, and the fixation prediction. We observe that the fixation masks effectively concentrate on the potential object while also accounting for their surroundings, mirroring human perception and highlighting the accuracy of our fixation prediction module. The subsequent graph illustrates the predicted influence of various attributes on the success of camouflage, with blue, green, and red denoting Surrounding Factors, Camouflaged Object-Self Factors, and Imaging Quality Factors, respectively. In \cref{fig:intermediate_output_a}, a hare camouflaged against a complex background of weed occlusion and shadow interference is shown. Here, Surrounding Factors (0.62) are the primary determinants of camouflage effectiveness, with Environmental Pattern Matching, Color Matching, and Environmental Shading being pivotal attributes, aligning with human perceptual insights. In contrast, \cref{fig:intermediate_output_b} depicts a seahorse seamlessly integrated into its coral environment through color and shape mimicry. Our model identifies Camouflaged Object-Self Factors (0.54) as the leading influence, highlighting shape mimicry as a crucial element, in agreement with human cognitive expectations. With these illustrations, the effectiveness of the proposed components is further demonstrated.

    \subsection{Discussion}

\begin{figure}[tb]
    \centering
      \begin{subfigure}{0.33\linewidth}
        \centering
        \includegraphics[height=2.7cm]{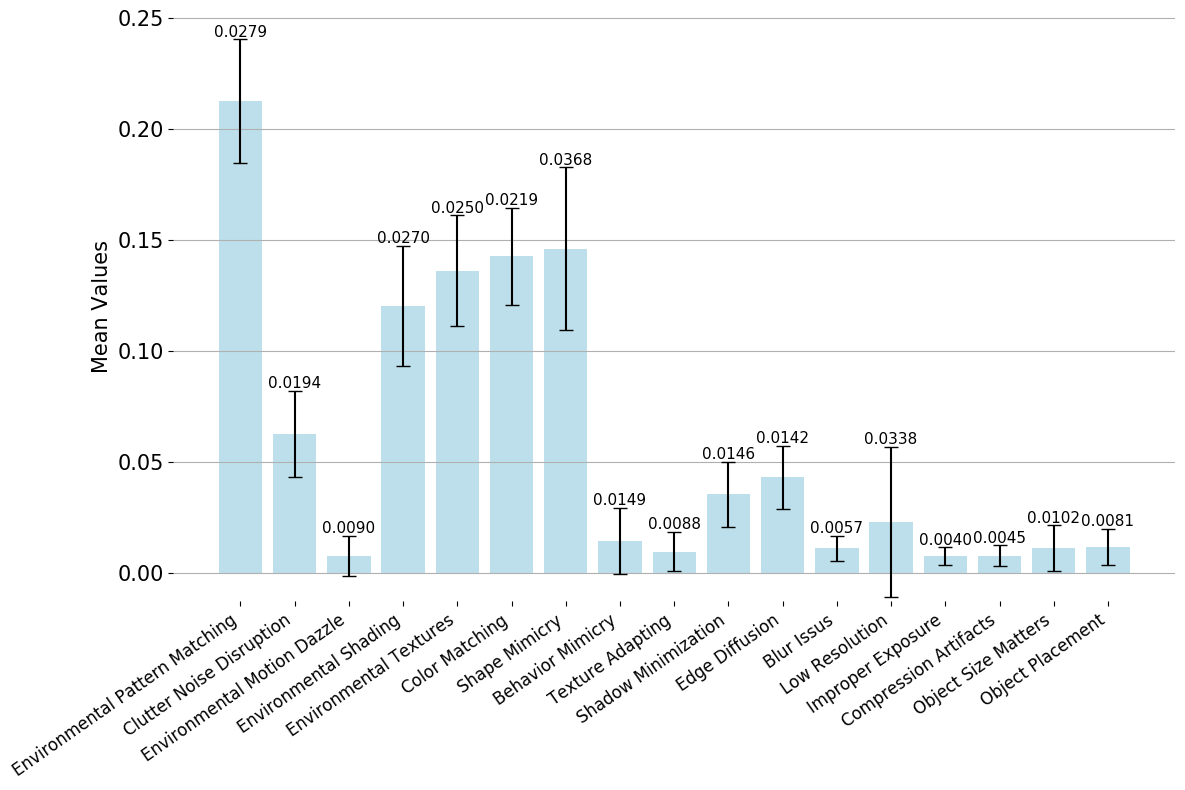}
        \caption{CAMO}
        \label{fig:discussion_CAMO}
      \end{subfigure}%
      \hfill
      \begin{subfigure}{0.33\linewidth}
        \centering
        \includegraphics[height=2.7cm]{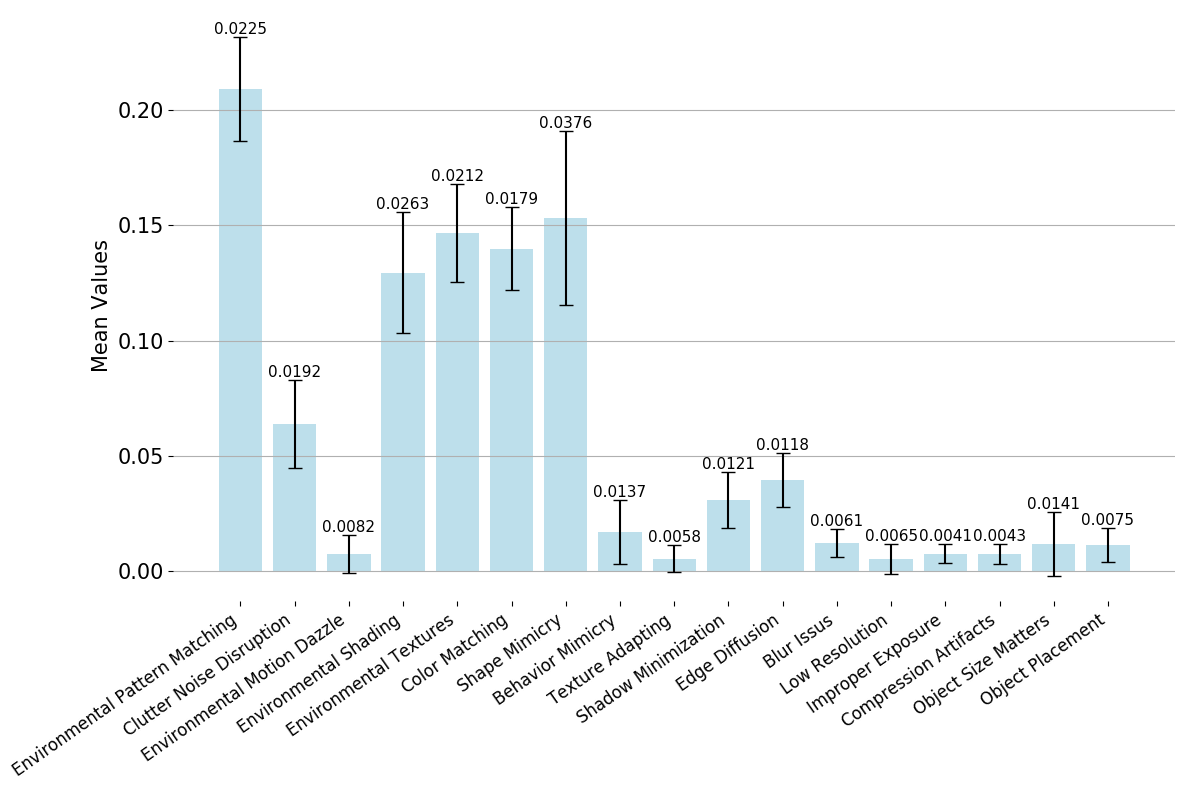}
        \caption{COD10K}
        \label{fig:discussion_COD10K}
      \end{subfigure}%
      \hfill
      \begin{subfigure}{0.33\linewidth}
        \centering
        \includegraphics[height=2.7cm]{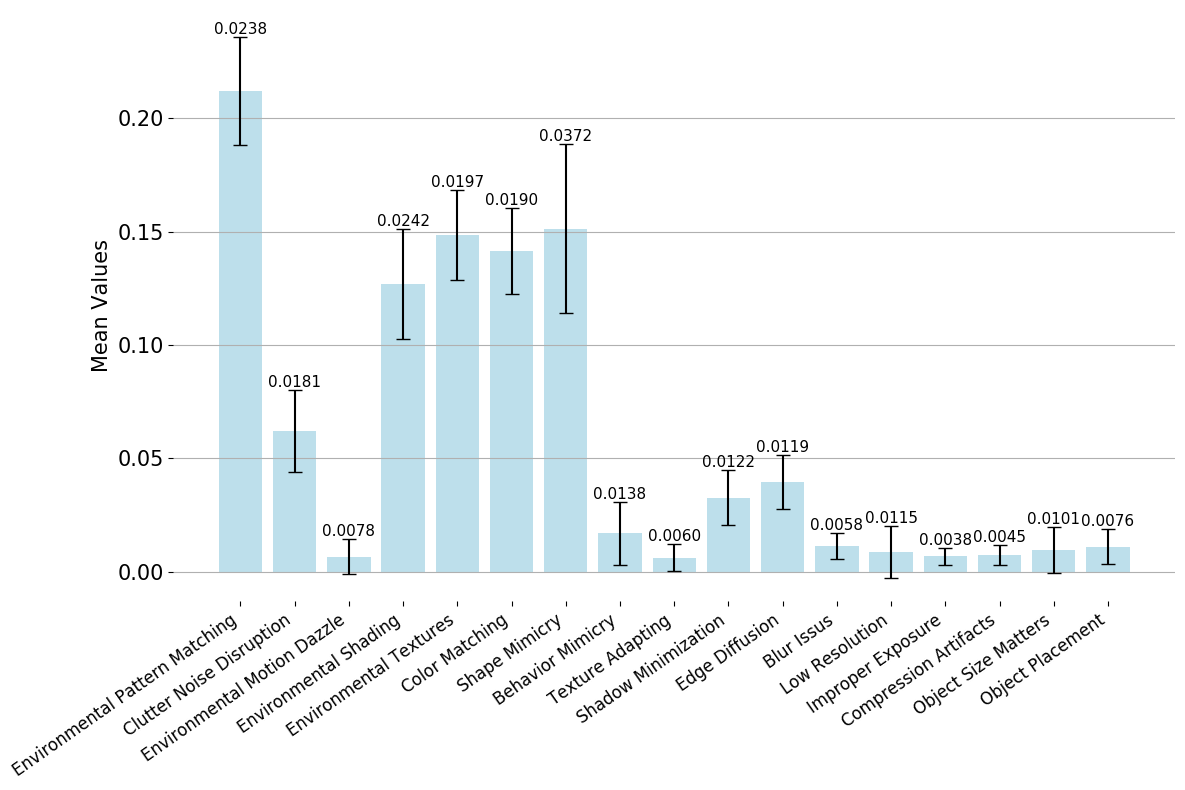}
        \caption{NC4K}
        \label{fig:discussion_NC4K}
      \end{subfigure}
      
      \caption{Attributes Contribution Statistic results. (Zoom in for details)}
      \label{fig:distribution_of_datasets}
    \end{figure}

    \textbf{Attributions' Contribution Among Different Datasets.} In \cref{fig:distribution_of_datasets}, we present an analysis of camouflage patterns across various testing datasets, using histogram bars to represent the proportional contribution of our proposed attributes and error bars to indicate the standard deviation. A comparison of mean values reveals that the COD10K and NC4K datasets predominantly feature Environmental Pattern Matching, Shape Mimicry, and Environmental Textures, which together account for over 50\% of their camouflage effectiveness. These attributes are crucial across all datasets, though their contributions vary. For instance, Shape Mimicry accounts for more than 15\% of camouflage success in both COD10K and NC4K, but less than 15\% in CAMO. We also notice that the mean distribution patterns of COD10K and NC4K are remarkably similar, reflecting their large and comparable sample sizes (2026 and 4040, respectively) that exhibit consistent camouflage patterns. In contrast, the CAMO dataset, with only 250 images, shows a mean distribution more prone to anomalies. For example, the standard deviation for Environmental Pattern Matching in CAMO is 0.0279, significantly higher than in COD10K (0.0225) and NC4K (0.0238). Additionally, the Low Resolution attribute in CAMO has a notably higher mean and standard deviation, indicating a higher prevalence of low-resolution images, which likely affects its MSE
    performance (0.0389) compared to COD10K (0.02592) and NC4K (0.03592) using same inference method. This trend is corroborated by other methodologies as shown in \cref{tab:quant_result}, further substantiating our findings. Further discussion including failure cases analysis could be found in Supplemental Materials.

\section{Conclusion}
In this paper, we presented a pioneering study on the role of camouflage attributes in determining the effectiveness of camouflage patterns, alongside the introduction of the COD-TAX dataset for comprehensive analysis. We also introduce the ACUMEN framework, which uniquely integrates textual and visual data for enhancing COS performance. Our findings, underscored by ACUMEN's superior performance over existing methods, highlight the significance of attribute analysis in camouflage designing and breaking. 

\textbf{Acknowledgement.} This work was supported by the National Natural Science Foundation of China (Grant No. 62002005); in part through the NYU IT High Performance Computing resources, services, and staff expertise.
%
%
\bibliographystyle{splncs04}
\bibliography{main}

\begin{thebibliography}{10}
\providecommand{\url}[1]{\texttt{#1}}
\providecommand{\urlprefix}{URL }
\providecommand{\doi}[1]{https://doi.org/#1}

\bibitem{bi2021rethinking}
Bi, H., Zhang, C., Wang, K., Tong, J., Zheng, F.: Rethinking camouflaged object detection: Models and datasets. IEEE TCSVT  \textbf{32}(9),  5708--5724 (2021)

\bibitem{chen2022boundary}
Chen, T., Xiao, J., Hu, X., Zhang, G., Wang, S.: Boundary-guided network for camouflaged object detection. Knowledge-Based Systems  \textbf{248},  108901 (2022)

\bibitem{cheng2023large}
Cheng, S., Ji, G.P., Qin, P., Fan, D.P., Zhou, B., Xu, P.: Large model based referring camouflaged object detection. arXiv  (2023)

\bibitem{cong2023frequency}
Cong, R., Sun, M., Zhang, S., Zhou, X., Zhang, W., Zhao, Y.: Frequency perception network for camouflaged object detection. In: ACM MM. pp. 1179--1189 (2023)

\bibitem{dosovitskiy2021an}
Dosovitskiy, A., Beyer, L., Kolesnikov, A., Weissenborn, D., Zhai, X., Unterthiner, T., Dehghani, M., Minderer, M., Heigold, G., Gelly, S., Uszkoreit, J., Houlsby, N.: An image is worth 16x16 words: Transformers for image recognition at scale. In: ICLR (2021), \url{https://openreview.net/forum?id=YicbFdNTTy}

\bibitem{fan2017structure}
Fan, D.P., Cheng, M.M., Liu, Y., Li, T., Borji, A.: Structure-measure: A new way to evaluate foreground maps. In: IEEE ICCV. pp. 4548--4557 (2017)

\bibitem{fan2018enhanced}
Fan, D.P., Gong, C., Cao, Y., Ren, B., Cheng, M.M., Borji, A.: Enhanced-alignment measure for binary foreground map evaluation. In: IJCAI. pp. 698--704 (2018)

\bibitem{fan2021concealed}
Fan, D.P., Ji, G.P., Cheng, M.M., Shao, L.: Concealed object detection. IEEE TPAMI  \textbf{44}(10),  6024--6042 (2021)

\bibitem{fan2020camouflaged}
Fan, D.P., Ji, G.P., Sun, G., Cheng, M.M., Shen, J., Shao, L.: Camouflaged object detection. In: IEEE CVPR. pp. 2777--2787 (2020)

\bibitem{fan2023advances}
Fan, D.P., Ji, G.P., Xu, P., Cheng, M.M., Sakaridis, C., Van~Gool, L.: Advances in deep concealed scene understanding. Visual Intelligence  \textbf{1}(1), ~16 (2023)

\bibitem{fan2020pranet}
Fan, D.P., Ji, G.P., Zhou, T., Chen, G., Fu, H., Shen, J., Shao, L.: Pranet: Parallel reverse attention network for polyp segmentation. In: MICCAI. pp. 263--273 (2020)

\bibitem{he2023camouflaged}
He, C., Li, K., Zhang, Y., Tang, L., Zhang, Y., Guo, Z., Li, X.: Camouflaged object detection with feature decomposition and edge reconstruction. In: IEEE CVPR. pp. 22046--22055 (2023)

\bibitem{he2023clip}
He, W., Jamonnak, S., Gou, L., Ren, L.: Clip-s4: Language-guided self-supervised semantic segmentation. In: IEEE CVPR. pp. 11207--11216 (2023)

\bibitem{hou2011detection}
Hou, J.Y.Y.H.W., Li, J.: Detection of the mobile object with camouflage color under dynamic background based on optical flow. Procedia Engineering  \textbf{15},  2201--2205 (2011)

\bibitem{hu2023relax}
Hu, J., Lin, J., Cai, W., Gong, S.: Relax image-specific prompt requirement in sam: A single generic prompt for segmenting camouflaged objects. arXiv  (2023)

\bibitem{hu2018squeeze}
Hu, J., Shen, L., Sun, G.: Squeeze-and-excitation networks. In: IEEE CVPR. pp. 7132--7141 (2018)

\bibitem{hu2023high}
Hu, X., Wang, S., Qin, X., Dai, H., Ren, W., Luo, D., Tai, Y., Shao, L.: High-resolution iterative feedback network for camouflaged object detection. In: AAAI. vol.~37, pp. 881--889 (2023)

\bibitem{huang2023feature}
Huang, Z., Dai, H., Xiang, T.Z., Wang, S., Chen, H.X., Qin, J., Xiong, H.: Feature shrinkage pyramid for camouflaged object detection with transformers. In: IEEE CVPR. pp. 5557--5566 (2023)

\bibitem{ji2023deep}
Ji, G.P., Fan, D.P., Chou, Y.C., Dai, D., Liniger, A., Van~Gool, L.: Deep gradient learning for efficient camouflaged object detection. Machine Intelligence Research  \textbf{20}(1),  92--108 (2023)

\bibitem{kingma2014adam}
Kingma, D.P., Ba, J.: Adam: A method for stochastic optimization. ICLR  (2015)

\bibitem{le2019anabranch}
Le, T.N., Nguyen, T.V., Nie, Z., Tran, M.T., Sugimoto, A.: Anabranch network for camouflaged object segmentation. CVIU  \textbf{184},  45--56 (2019)

\bibitem{li2021uncertainty}
Li, A., Zhang, J., Lv, Y., Liu, B., Zhang, T., Dai, Y.: Uncertainty-aware joint salient object and camouflaged object detection. In: IEEE CVPR. pp. 10071--10081 (2021)

\bibitem{li2023blip}
Li, J., Li, D., Savarese, S., Hoi, S.: Blip-2: Bootstrapping language-image pre-training with frozen image encoders and large language models. arXiv  (2023)

\bibitem{li2022grounded}
Li, L.H., Zhang, P., Zhang, H., Yang, J., Li, C., Zhong, Y., Wang, L., Yuan, L., Zhang, L., Hwang, J.N., et~al.: Grounded language-image pre-training. In: IEEE CVPR. pp. 10965--10975 (2022)

\bibitem{liu2024visual}
Liu, H., Li, C., Wu, Q., Lee, Y.J.: Visual instruction tuning. NeurIPS  \textbf{36} (2024)

\bibitem{liu2023explicit}
Liu, W., Shen, X., Pun, C.M., Cun, X.: Explicit visual prompting for low-level structure segmentations. In: IEEE CVPR. pp. 19434--19445 (2023)

\bibitem{liu2012foreground}
Liu, Z., Huang, K., Tan, T.: Foreground object detection using top-down information based on em framework. IEEE TIP  \textbf{21}(9),  4204--4217 (2012)

\bibitem{lv2023towards}
Lv, Y., Zhang, J., Dai, Y., Li, A., Barnes, N., Fan, D.P.: Towards deeper understanding of camouflaged object detection. IEEE TCSVT  (2023)

\bibitem{lv2021simultaneously}
Lv, Y., Zhang, J., Dai, Y., Li, A., Liu, B., Barnes, N., Fan, D.P.: Simultaneously localize, segment and rank the camouflaged objects. In: IEEE CVPR. pp. 11591--11601 (2021)

\bibitem{lyu2023uedg}
Lyu, Y., Zhang, H., Li, Y., Liu, H., Yang, Y., Yuan, D.: Uedg: Uncertainty-edge dual guided camouflage object detection. IEEE TMM  (2023)

\bibitem{lyu2023deltaedit}
Lyu, Y., Lin, T., Li, F., He, D., Dong, J., Tan, T.: Deltaedit: Exploring text-free training for text-driven image manipulation. In: IEEE CVPR. pp. 6894--6903 (2023)

\bibitem{margolin2014evaluate}
Margolin, R., Zelnik-Manor, L., Tal, A.: How to evaluate foreground maps? In: IEEE CVPR. pp. 248--255 (2014)

\bibitem{mayer2002multimedia}
Mayer, R.E.: Multimedia learning. In: Psychology of learning and motivation, vol.~41, pp. 85--139 (2002)

\bibitem{mei2021camouflaged}
Mei, H., Ji, G.P., Wei, Z., Yang, X., Wei, X., Fan, D.P.: Camouflaged object segmentation with distraction mining. In: IEEE CVPR. pp. 8772--8781 (2021)

\bibitem{merilaita2017camouflage}
Merilaita, S., Scott-Samuel, N.E., Cuthill, I.C.: How camouflage works. Philosophical Transactions of the Royal Society B: Biological Sciences  \textbf{372}(1724),  20160341 (2017)

\bibitem{paivio2013imagery}
Paivio, A.: Imagery and verbal processes. Psychology Press (2013)

\bibitem{pang2022zoom}
Pang, Y., Zhao, X., Xiang, T.Z., Zhang, L., Lu, H.: Zoom in and out: A mixed-scale triplet network for camouflaged object detection. In: IEEE CVPR. pp. 2160--2170 (2022)

\bibitem{pembury2020camouflage}
Pembury~Smith, M.Q., Ruxton, G.D.: Camouflage in predators. Biological Reviews  \textbf{95}(5),  1325--1340 (2020)

\bibitem{radford2021learning}
Radford, A., Kim, J.W., Hallacy, C., Ramesh, A., Goh, G., Agarwal, S., Sastry, G., Askell, A., Mishkin, P., Clark, J., et~al.: Learning transferable visual models from natural language supervision. In: ICML. pp. 8748--8763 (2021)

\bibitem{CognitionJohn}
Skelhorn, J., Rowe, C.: Cognition and the evolution of camouflage. Proceedings of the Royal Society B: Biological Sciences  \textbf{283}(1825),  20152890 (2016). \doi{10.1098/rspb.2015.2890}

\bibitem{song2023rinet}
Song, Y., Liu, Z., Li, G., Zeng, D., Zhang, T., Xu, L., Wang, J.: Rinet: Relative importance-aware network for fixation prediction. IEEE TMM  \textbf{25},  9263--9277 (2023)

\bibitem{stevens2009animal}
Stevens, M., Merilaita, S.: Animal camouflage: current issues and new perspectives. Philosophical Transactions of the Royal Society B: Biological Sciences  \textbf{364}(1516),  423--427 (2009)

\bibitem{stevens2019key}
Stevens, M., Ruxton, G.D.: The key role of behaviour in animal camouflage. Biological Reviews  \textbf{94}(1),  116--134 (2019)

\bibitem{tankus2001convexity}
Tankus, A., Yeshurun, Y.: Convexity-based visual camouflage breaking. CVIU  \textbf{82}(3),  208--237 (2001)

\bibitem{troscianko2009camouflage}
Troscianko, T., Benton, C.P., Lovell, P.G., Tolhurst, D.J., Pizlo, Z.: Camouflage and visual perception. Philosophical Transactions of the Royal Society B: Biological Sciences  \textbf{364}(1516),  449--461 (2009)

\bibitem{wei2020f3net}
Wei, J., Wang, S., Huang, Q.: F$^3$net: fusion, feedback and focus for salient object detection. In: AAAI. vol.~34, pp. 12321--12328 (2020)

\bibitem{wu2023source}
Wu, Z., Paudel, D.P., Fan, D.P., Wang, J., Wang, S., Demonceaux, C., Timofte, R., Van~Gool, L.: Source-free depth for object pop-out. In: IEEE CVPR. pp. 1032--1042 (2023)

\bibitem{xu2023side}
Xu, M., Zhang, Z., Wei, F., Hu, H., Bai, X.: Side adapter network for open-vocabulary semantic segmentation. In: IEEE CVPR. pp. 2945--2954 (2023)

\bibitem{yan2023camouflaged}
Yan, X., Sun, M., Han, Y., Wang, Z.: Camouflaged object segmentation based on matching--recognition--refinement network. IEEE TNNLS  (2023)

\bibitem{yang2023implicit}
Yang, S., Ding, M., Wu, Y., Li, Z., Zhang, J.: Implicit neural representation for cooperative low-light image enhancement. In: IEEE ICCV. pp. 12918--12927 (2023)

\bibitem{zhang2022glipv2}
Zhang, H., Zhang, P., Hu, X., Chen, Y.C., Li, L., Dai, X., Wang, L., Yuan, L., Hwang, J.N., Gao, J.: Glipv2: Unifying localization and vision-language understanding. NeurIPS  \textbf{35},  36067--36080 (2022)

\bibitem{zhang2023cfanet}
Zhang, Q., Yan, W.: Cfanet: A cross-layer feature aggregation network for camouflaged object detection. In: IEEE ICME. pp. 2441--2446 (2023)

\bibitem{zhang2023referring}
Zhang, X., Yin, B., Lin, Z., Hou, Q., Fan, D.P., Cheng, M.M.: Referring camouflaged object detection. arXiv  (2023)

\bibitem{zheng2023mffn}
Zheng, D., Zheng, X., Yang, L.T., Gao, Y., Zhu, C., Ruan, Y.: Mffn: Multi-view feature fusion network for camouflaged object detection. In: IEEE WACV. pp. 6232--6242 (2023)

\end{thebibliography}
\end{document}